\documentclass[runningheads]{llncs}

\usepackage{eccvabbrv}

\usepackage{graphicx}
\usepackage{booktabs}
\usepackage{xcolor}
\usepackage{multirow}

\usepackage[accsupp]{axessibility}  

\usepackage{hyperref}

\usepackage{orcidlink}

\begin{document}

\title{\textcolor{red}{O}\textcolor{purple}{S}\textcolor{green}{T}\textcolor{blue}{A}\textcolor{brown}{F}: A \textcolor{red}{O}ne-\textcolor{purple}{S}hot \textcolor{green}{T}uning Method for \\Improved \textcolor{blue}{A}ttribute-\textcolor{brown}{F}ocused T2I Personalization} 

\titlerunning{Attribute-Aware Customization}
\author{Ye Wang\inst{1} \hspace{0.5cm} Zili Yi\inst{2} \hspace{0.4cm}  Rui Ma\inst{1,3,\thanks{Corresponding author}}}

\authorrunning{Ye Wang et al.}

\institute{School of Artificial Intelligence, Jilin University, Changchun, China \and
School of Intelligence Science and Technology, Nanjing University, Suzhou, China \and
Engineering Research Center of Knowledge-Driven Human-Machine Intelligence, MOE, China
\\
}

\maketitle

\begin{figure}
    \centering
    \includegraphics[width=\linewidth]{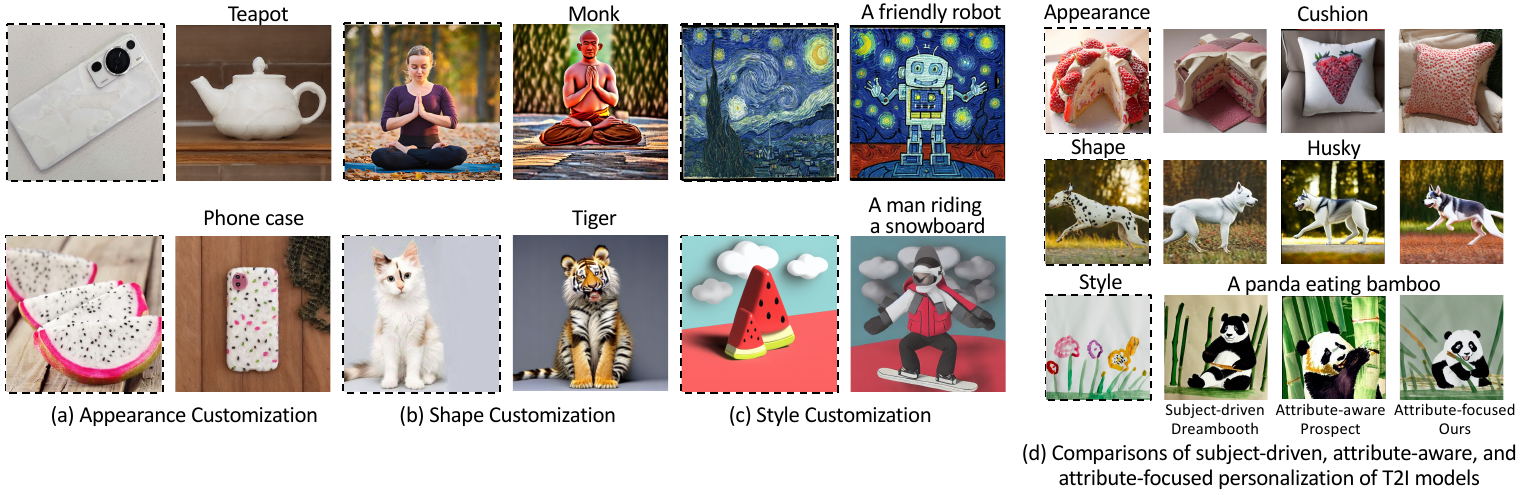}
    \caption{Attribute-focused text-to-image personalization. Our method allows for the generation of customized appearance, shape and style attributes using only one reference image, as shown by the dashed frame.}
    \label{fig:teaser}
  \end{figure}

\vspace{-30pt}

\begin{abstract}
Personalized text-to-image (T2I) models not only produce lifelike and varied visuals but also allow users to tailor the images to fit their personal taste. These personalization techniques can grasp the essence of a concept through a collection of images, or adjust a pre-trained text-to-image model with a specific image input for subject-driven or attribute-aware guidance. Yet, accurately capturing the distinct visual attributes of an individual image poses a challenge for these methods. To address this issue, we introduce OSTAF, a novel parameter-efficient one-shot fine-tuning method which only utilizes one reference image for T2I personalization.
A novel hypernetwork-powered attribute-focused fine-tuning mechanism is employed to achieve the precise learning of various attribute features (e.g., appearance, shape or drawing style) from the reference image.
Comparing to existing image customization methods, our method shows significant superiority in attribute identification and application, as well as achieves a good balance between efficiency and output quality.
\end{abstract}

\section{Introduction}

Over the recent years, significant progress has been observed in the area of customized Text-to-Image (T2I) generation \cite{gal2022image,ruiz2023dreambooth,kumari2023multi,han2023svdiff,gal2023designing,wei2023elite,ruiz2023hyperdreambooth,chen2023subject,shi2023instantbooth,jia2023taming,avrahami2023break,xiao2023fastcomposer,gal2023encoder}. These advancements in generative models have not only facilitated the generation of images that are both realistic and varied, but have also empowered users to shape these images to align with their personal visual preferences, i.e., image personalization. The latest methods in personalization have been adept at extracting the essence of a unified concept from an set of images, subsequently employing text-based prompts to create new images that embodies these subjects \cite{ruiz2023dreambooth,kumari2023multi,gal2022image,shi2023instantbooth}. Furthermore, some approaches have pioneered the integration of specific imagery subjects into an pre-trained text-to-image diffusion model by employing an image encoder, thus enabling one-shot subject-aware personalization \cite{wei2023elite,gal2023encoder,sohn2023styledrop}.

Specifically, Gal et al. \cite{gal2022image} observe that the shallow layers of the denoising U-net structures within diffusion models tend to generate colors and materials, while the deep layers provide semantic guidance. They use only 3-5 images to learn a user-provided concept and represent it using new “words” in the embedding space of a frozen text-to-image model. P+ \cite{voynov2023p+} extends a single text prompt into multiple prompts and injects them into different cross-attention layers of U-net to decouple visual attributes like style, color and structure. However, P+ \cite{voynov2023p+} requires multiple reference images of a specific subject, which can be hard to collect. On the other hand, StyleDrop \cite{sohn2023styledrop} allows one-shot style-aware or content-aware personalization of text-to-image synthesis, by tuning a specific subset of parameters of a text-to-image diffusing model. Zhang et al. \cite{zhang2023prospect} introduce an expanded conditioning space to enable the representation, disentanglement and recombination of visual attributes of a given exemplar image to guide image synthesis and editing.

Nonetheless, these methods for customized T2I synthesis, whether targeted on subject-aware or attribute-aware personalization, often struggle with or fall short in precisely separating attribute features: see Figure \ref{fig:teaser} (d). Capturing the distinct and standout visual characteristics of a single reference image, which we denote as \textit{attribute-focused personalization}, continues to be a challenge. For instance, individuals might want to personalize specific visual attributes, like creating a phone case that mimics the dominant material seen in the reference image or illustrating a tiger in a pose reminiscent of a cat's seated position as depicted in the reference, as shown in Figure \ref{fig:teaser} (a) and (b). Achieving these effects demands the precise identification and disentanglement of the distinct and notable visual attributes of the reference image, which is beyond the vague capture of blended image-level features or subjects.

In this paper, we propose OSTAF, a straightforward yet effective hypernetwork-enhanced one-shot fine-tuning method for attribute-focused T2I personalization. 
Our goal is to achieving efficient and high-quality attribute-focused (e.g., appearance, shape and style) personalization by fine-tuning pre-trained text-to-image diffusion models with only a single reference image, as shown in Figure \ref{fig:teaser} (a), (b) and (c). 
Initially, we examine how the parameters of the U-net encoder and decoder in the diffusion model process various visual attributes. We discovered that the encoder primarily learns shapes, while the decoder is more attuned to appearance and style. A naive strategy might involve fine-tuning only the relevant U-net parameters for the desired visual attribute. However, this method is flawed, as relying solely on a single reference image can lead to catastrophic overfitting, significantly impairing the model's ability to control the image generation based on text. To address these issues, we employ an additional lightweight hypernetwork to modulate and refine the U-net's weights, as used in \cite{gal2023encoder,alaluf2022hyperstyle}, while we focus on tuning only the encoder or decoder instead of the whole network for learning an certain attribute. 
This strategy not only ensures smoother updates of the parameters and reduces the likelihood of overfitting, but also can effectively identify and represent the desired attributes from the reference image.


For evaluation, we compare with existing approaches on a dataset collected for attribute-level customization. 
From the quantitative and qualitative results, our method demonstrates superiority in the customization of various attributes. 
Furthermore, our method can also achieve a good balance between efficiency and output quality.

In summary, the key contributions of our work are outlined as follows:
\begin{itemize}
\item We introduce OSTAF, a streamlined and remarkably efficient one-shot fine-tuning approach, which only utilizes a single reference image, for attribute-focused text-to-image personalization.
\item We delineate the distinct roles of the encoder and decoder of the diffusion U-net framework in accurately capturing diverse attributes, such as appearance, shape and style, and employ a lightweight hypernetwork for more focused attribute identification and application.

\item Through comprehensive quantitative and qualitative evaluation, we show that our method significantly outperforms existing image customization methods for attribute identification and attribute-focused personalization.
\end{itemize}

\section{Related Works}
\label{sec:related_works}

\paragraph{\textbf{Personalized Text-to-Image Generation.}} Recent studies \cite{sun2023imagebrush,nguyen2023visual,bar2022visual,tumanyan2022splicing,yang2023paint,xu2023prompt,goel2023pair} have pivoted towards using visual exemplars as a cornerstone for image generation to navigate the inherent vagueness and unpredictability associated with text-based prompts. This methodology emphasizes the use of one or more reference images as a primary guide, moving away from the exclusive dependence on textual descriptions for synthesizing images.
Nonetheless, these approaches tend to concentrate on capturing the general essence of the reference image, such as its objects or subjects, without the capacity for attribute-focused text-to-image (T2I) customization. Furthermore, several methodologies \cite{yang2023paint,goel2023pair} are characterized by their substantial training demands, requiring extensive fine-tuning across vast datasets to enable the use of visual images as conditional inputs for Stable Diffusion. In contrast, our proposed method seeks to overcome the limitations associated with extracting multiple visual attributes from a singular image. It encompasses the processes of identifying, separating, and reassembling visual attributes, offering a nuanced approach to T2I personalization.


\paragraph{\textbf{Parameter Efficient Fine Tuning (PEFT).}} PEFT represents an innovative approach in the refinement of deep learning models, emphasizing the adjustment of a subset of parameters rather than the entire model. These parameters are identified as either specific subsets from the originally trained model or a minimal number of newly introduced parameters during the fine-tuning phase. PEFT has been applied in text-to-image diffusion models \cite{saharia2022photorealistic,rombach2022high} through techniques such as LoRA \cite{ryu2023low} and adapter tuning \cite{mou2023t2i,ye2023ip,wei2023elite,chen2024subject,ma2023unified}.
To facilitate the adaptation of pre-trained text-to-image generators to visual inputs, ELITE \cite{wei2023elite} fine-tunes the attention layer parameters, while UMM-Diffusion \cite{ma2023unified} introduces a visual mapping layer, keeping the pre-trained generator's weights unchanged. SuTI \cite{chen2024subject} enables personalized image generation without the need for test-time fine-tuning by leveraging a vast dataset of images created by subject-specific expert models. Contrary to these approaches, our method leverages a hypernetwork framework to adjust and refine a unique subset of pre-trained parameters.

\paragraph{\textbf{Many-shot T2I Personalization.}} 
Many-shot techniques \cite{bansal2023universal,valevski2023face0,yuan2023inserting} necessitate the training of either the diffusion model itself or its conditioning branch to facilitate customized text-to-image (T2I) generation, relying on extensive datasets or a handful of examples for training. DreamBooth \cite{ruiz2023dreambooth} introduces a methodology for embedding a new subject into the existing model architecture without compromising the model's original capabilities, by training the diffusion model with reference samples. In contrast, SuTI \cite{chen2024subject} begins by assembling a substantial dataset of input images paired with their recontextualized counterparts generated via the standard DreamBooth procedure. InstantBooth \cite{shi2023instantbooth} devises a novel conditioning branch within the diffusion model, enabling personalization with a limited set of images to produce tailored outputs across various styles. 
FastComposer \cite{xiao2023fastcomposer} employs an image encoder to derive subject-specific embeddings, addressing the challenge of identity preservation when generating images with multiple subjects. 
Diverging from these many-shot strategies, our research concentrates on achieving T2I personalization with a one-shot approach.

\section{Method}

Given a single reference image, our goal is to distinguish, separate and learn different visual attributes, including appearance, shape, and style, and to facilitate the generation of attribute-focused text-to-image customization. To achieve this goal, we propose OSTAF, as illustrated in Figure \ref{fig:pipeline}.



In the following sections, we begin by introducing the preliminary of Stable Diffusion\cite{rombach2022high} in section \ref{subsec:Preliminary}. 
Subsequently, we present the learning preferences analysis of the Diffusion U-net encoder and decoder for different visual attributes in Section \ref{section:attr_analysis}. Then we introduce our framework in Section \ref{subsec:Architecture}, and finally, demonstrate the implementation details in Section \ref{subsec:Implementation}.

\subsection{Preliminary}
\label{subsec:Preliminary}

\textbf{Stable Diffusion.} 
Stable Diffusion\cite{rombach2022high}, a state-of-the-art text-to-image generation model, operates within a low-dimensional latent space. It begins by encoding an input image $x$ into a latent representation $z$ using a VAE encoder. Noise $\epsilon$ is then introduced at time step $t$ to create a noisy latent $z_t$. To guide the generation process with text conditions, Stable Diffusion incorporates a CLIP text encoder $\tau$ to encode textual prompts $c$, which are integrated into the cross-attention layers for interaction with the noisy latents. Finally, a conditional U-Net backbone $\epsilon_\theta$ is trained to predict the noise $\epsilon$. The training objectives is as follows:

\begin{equation}
    L_{SD}(\theta) := \mathbb{E}_{t,x_0,\epsilon} \left[ \lVert \epsilon - \epsilon_\theta(z_t, t, \tau(c)) \rVert^2 \right] \label{eq:LLDM}
\end{equation}

\subsection{Attribute Learning Preferences Analysis}
\label{section:attr_analysis}
The representation spaces associated with different visual attributes emphasize information and features at specific levels. The shape attribute predominantly captures low-level visual features, the appearance attribute focuses on intricate details such as texture, color, and material, and the style attribute reflects the overall stylistic characteristics of an image. Consequently, these attributes necessitate specialized learning within different modules of the network.

\begin{figure}
    \centering
    \includegraphics[width=1.0\textwidth]{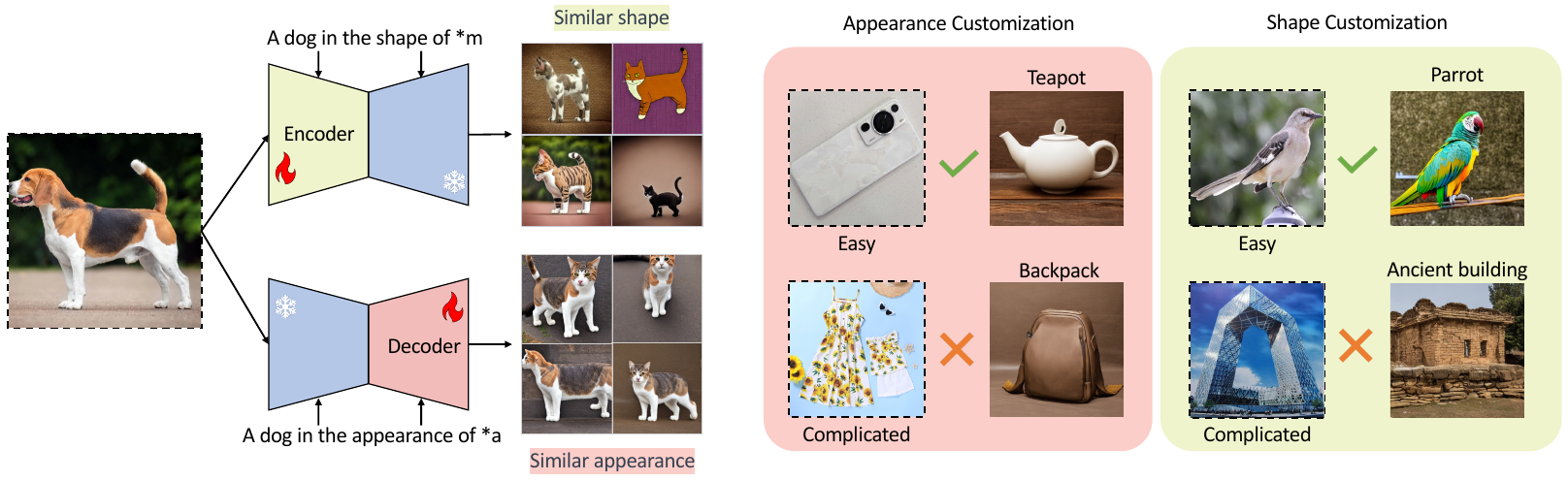}
    \caption{(Left) Illustration showcasing the unique roles of the encoder and decoder within the diffusion U-Net in learning varying attributes. (Right) A display of the outcomes achieved without tuning the hypernetwork, suggesting its effectiveness is limited to simple iconic reference images.}
    \label{fig:easy_hard}
\end{figure}

Based on the above considerations, we conducted a simple experiment based on Stable Diffusion\cite{rombach2022high}. As shown in Figure \ref{fig:easy_hard} left, we selected a reference image, for example, a dog, and inputted default text prompts such as "a dog in the appearance/shape of *a/*m." Subsequently, we separately fine-tuned the encoder and decoder modules of the U-net and utilized the fine-tuned model to generate the corresponding images. The text prompt used for inference is "a cat in the appearance/shape of *a/*m". The *a and *m are learnable token embeddings.

We observe that there is a significant difference in the generated results when separately fine-tuning the encoder and decoder. When fine-tuning the encoder, the generated images exhibit similar shape to reference dog, whereas fine-tuning the decoder leads to images with similar appearance. This experiment further validates our idea that different visual attributes are learned by distinct network modules.

However, we find that the above conclusion only works for simple images (pure color appearance or simple shape) and may not effectively accomplish attribute-focused customization for more complex images, as illustrated in Figure \ref{fig:easy_hard} right. This could possibly be due to catastrophic overfitting caused by fine-tuning the network on a single complex image. To address this issue, we propose a efficient hypernetwork-driven fine-tuning method to achieve smoother parameter updates to alleviate the risk of overfitting. In the following, we will introduce our overall framework and the details of our method.

\subsection{OSTAF Framework}
\label{subsec:Architecture}

\begin{figure*}
    \centering
    \includegraphics[width=1.0\textwidth]{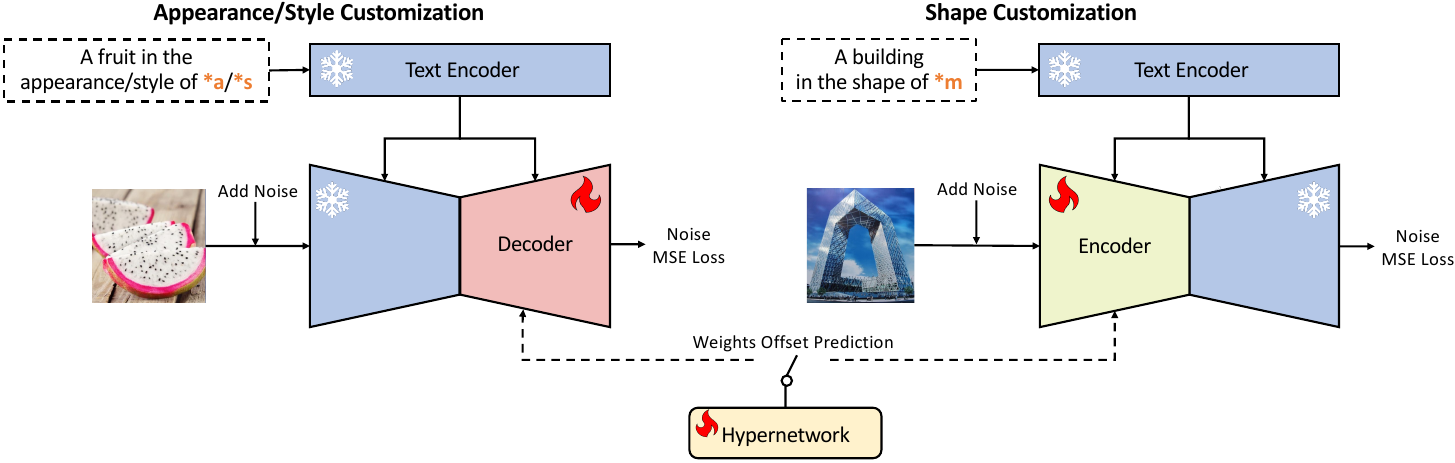}
    \caption{OSTAF pipeline. Our method requires only one reference image as input, and we introduce a hypernetwork-driven fine-tuning approach to adjust the parameters of the U-net encoder or decoder for efficient attribute-focused T2I customization.}
    \label{fig:pipeline}
\end{figure*}

As shown in Figure \ref{fig:pipeline}, our method architecture is streamlined and remarkably efficient, with its core component being a lightweight hypernetwork. We only need to input a single reference image and corresponding text prompt to capture the target attribute. This significantly reduces the demand for a large number of reference images.

\textbf{Hypernetwork-Driven Fine-tuning.} Fine-tuning all parameters or specific parameters of the U-net in Stable Diffusion to achieve image customization generation is a common practice\cite{ruiz2023dreambooth,kumari2023multi}. 
However, such fine-tuning approaches is not efficient, it can easily lead to catastrophic overfitting when fine-tuning on a single reference image, thus failing to achieve attribute-focused customization. To address this issue, we propose a efficient hypernetwork-driven fine-tuning mechanism, where the core idea is to utilize a lightweight hypernetwork to modulate and guide the parameter updates of U-net's encoder or decoder, rather than performing direct fine-tuning. Essentially, the hypernetwork is trained to guide the updates of the main network parameters in a low-rank, smooth manner. The structure of the hypernetwork is highly lightweight, consisting only of four linear layers.  This efficient fine-tuning approach greatly reduces the risk of overfitting when fine-tuning on a single reference image, while also achieving high-quality, attribute-controllable customized generation.

\begin{figure}
    \centering
    \includegraphics[width=0.8\textwidth]{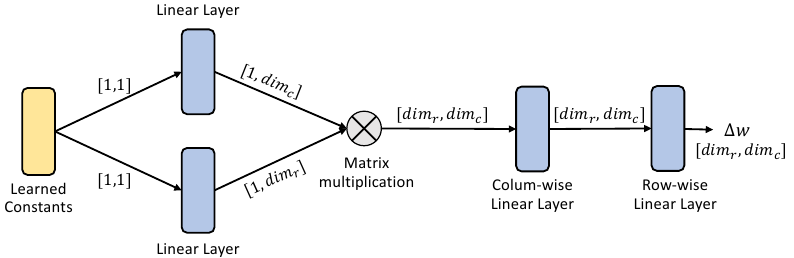}
    \caption{The architecture of hypernetwork.}
    \label{fig:hypernetwork}
\end{figure}

The hypernetwork architecture is  shown in the Figure \ref{fig:hypernetwork}, we follow the architecture of E4T\cite{gal2023encoder} weight offsets prediction module for the construction of our attribute hypernetworks. 
The module takes as input a learnable constant $cons$ (default-initialized to 1) and the dimension information $[dim_{r}, dim_{c}]$ of the target weight parameters. It is then trained to predict weight offsets in the same dimensions as the target weight parameters. Here, $dim_{r}$ represents the number of rows of the target weight parameters, and $dim_{c}$ represents the number of columns.
In detail, the learnable constant passes through two linear layers, yielding outputs that are multiplied to derive the initial weight offset matrix. Row and column transformations are then applied to this matrix to obtain the final weight offset matrix $\Delta w$. As discussed in the literatures\cite{gal2023encoder,kumari2023multi,wei2023elite}, the weights of self-attention and cross-attention play a crucial role in the process of image customization.
Therefore, we utilize the hypernetwork as a weight offsets prediction module to modulate and guide the updates of attention-related weights within the encoder or decoder.
The high-level parameter update process is defined as follows:

\begin{align}
    \Delta w &= hypernetwork(cons,dim_{r},dim_{c})  \\
    w^*_{attn} &= w_{attn} + \lambda * \Delta w
\end{align}
the $w_{attn}$ denotes the general term for the attention-related parameters, which includes the query matrix, key matrix and value matrix for self-attention and cross-attention layers. $\lambda$ is a weight coefficient that is used to regulate the updating strength of parameters. During training, we set $\lambda$ to 1.0. Once training is complete, during inference generation, we can adjust the value of $\lambda$ to control the strength of attribute customization.



\textbf{Loss Function.} To guide the customization and learning of attributes, we employ the original noise prediction loss function, which is expressed as:

\begin{equation}
    L_{OSTAF}(\theta) := \mathbb{E}_{t,x_0,\epsilon} \left[ \lVert \epsilon - \epsilon_\theta(x_t, t, \tau(c)) \rVert^2 \right] \label{eq:LLDM}
\end{equation}
Note that $\theta$ denotes the parameters of the encoder or decoder and the corresponding hypernetwork, $\epsilon$ denotes the noise, $z_t$ represents the latent of input image at time $t$, $t$ denotes the current time step, $\tau(c)$ represents the encoding of the input text prompt $c$ using the text encoder $\tau$ of the CLIP model.

\subsection{Implementation Details}
\label{subsec:Implementation} 
We employ Stable Diffusion 1.4\cite{rombach2022high} as our base model. During the training process, the visual encoder and text encoder are kept frozen. We only need a single reference image. The input text prompt is in the form of "a \textit{class name} in the shape/appearance/style of *m/*a/*s". For shape attribute customization, we solely apply the resize augmentation. For appearance and style attribute customization, we employ random cropping and horizontal flipping. To avoid background interference, we extract the foreground content using SAM for appearance and style customization. Our model is trained on a single NVIDIA A40 GPU with a batch size of 1 and a learning rate set to 1e-6. The fine-tuning steps and time for each reference image may vary slightly. On average, about 1000 iterations are required for attribute-focused customized generation, which typically takes around 10 minutes to complete, significantly less than the more than 20 minutes required for Prospect.

\section{Experiments}
\label{sec:experiments}

\subsection{Attribute Benchmark}
\label{sec:attribute_benchmark}

\textbf{Attribute Benchmark.} Presently, there exists a shortage of dedicated datasets for the evaluation of attribute-focused customization generation. Therefore, we collect and introduce a novel benchmark known as the "Attribute Benchmark". This benchmark consists of three sub-datasets: a shape dataset with 13 images, an appearance dataset with 12 images, and a style attribute dataset with 9 images. The style dataset has been curated from the images utilized in the StyleDrop\cite{sohn2023styledrop}.

\indent{\textbf{Evaluation Metrics.}}  As suggested by \cite{alaluf2023cross}, we employ CLIP-T and IoU scores for evaluating shape attribute customization, while for appearance and style attribute customization, we utilize CLIP-T score and Gram matrix distances. The CLIP-T score measures the similarity between the generated images and textual prompts, whereas the IoU score quantifies the shape consistency between binary masks extracted from generated images and reference images. The Gram matrix distance measures the stylistic similarity between images; the smaller the distance, the more similar the styles\cite{gatys2015neural}. Additionally, this metric can be employed to assess appearance similarity, as elucidated in \cite{alaluf2023cross}.
Furthermore, we incorporate the DINO similarity score to assess the consistency of appearance attributes between generated images and reference images using DINO CLS features\cite{goel2023pair}.

\indent{\textbf{Baseline and SOTA Methods.}} We performed a comparative analysis with the tuning-based approach and the adapter-based approach, respectively. The tuning-based methods comprise Dreambooth\cite{ruiz2023dreambooth}, Custom-Diffusion\cite{kumari2023multi}, and Prospect\cite{zhang2023prospect}. The adapter-based methods include IP-Adapter\cite{ye2023ip} and ControlNet\cite{zhang2023adding} (ControlNet-Canny, ControlNet-Segmentation).
Furthermore, we choose Stable Diffusion\cite{rombach2022high} as the baseline model.

\subsection{Quantitative Experiments}

\textbf{Shape and Appearance Customization Evaluation.} We individually trained the Stable Diffusion\cite{rombach2022high}, DreamBooth\cite{ruiz2023dreambooth}, Custom Diffusion\cite{kumari2023multi}, Prospect\cite{zhang2023prospect}, and our method on the Attribute Benchmark dataset.  For the Stable Diffusion\cite{rombach2022high} baseline, we have individually fine-tuned the encoder and decoder of the U-Net. For Dreambooth\cite{ruiz2023dreambooth} and Custom Diffusion\cite{kumari2023multi}, we have adapted the text prompts required during training to "a \textit{class name} in the shape/appearance of <shape>/<appearance>." for attribute learning. For Prospect\cite{zhang2023prospect}, we have adopted the same training and testing techniques as outlined in the original paper. For each reference image, we utilized approximately three distinct textual prompts, resulting in three generated images per text prompt. In total, each method underwent testing and produced 216 images, with 93 images dedicated to shape attribute and 123 images for appearance attribute.

\begin{table*}[]
    \caption{Quantitative comparison with respect to the tuning-based methods for shape and appearance attribute customization.}
    \label{tab:compare_sota}
    \resizebox{\linewidth}{!}{%
        \begin{tabular}{c|ccc|cccc}
            \hline
Metric                                                  & \multicolumn{3}{c|}{Appearance Customization}                                                        & \multicolumn{3}{c}{Shape Customization}                           \\ \hline
Method                                                  & CLIP-T  $\uparrow$ & DINO Similarity  $\uparrow$ & Gram Matrics $\downarrow$  & CLIP-T  $\uparrow$ & IoU Score $\uparrow$  \\ \hline
Stable Diffusion \cite{rombach2022high} & \textbf{0.2854}    & 0.2596                      & 0.1018                                        & 0.2380             & 0.3418                                   \\ \hline
DreamBooth \cite{ruiz2023dreambooth}    & 0.2681             & 0.2671                      & 0.0864                                        & 0.2791             & 0.3524                                  \\ \hline
Custom Diffusion \cite{kumari2023multi} & 0.2710             & 0.2977                      & 0.0817                                        & \textbf{0.2844}    & 0.3691                                   \\ \hline
Prospect \cite{zhang2023prospect}       & 0.2800             & 0.3849                      & 0.0842                                        & 0.2795             & 0.4656                                  \\ \hline
Ours                                    & 0.2822             & \textbf{0.4149}             & \textbf{0.0791}                               & 0.2798             & \textbf{0.4938}                        \\ \hline
        \end{tabular}%
    }
\end{table*}

\textbf{Comparison with Tuning-based Methods.} As shown in Table \ref{tab:compare_sota}, the Stable Diffusion\cite{rombach2022high} exhibits poor performance in the attributes customization. This indicates that a simple fine-tuning of the U-net encoder and decoder is insufficient for achieving precise attribute-focused customization. Dreambooth \cite{ruiz2023dreambooth} and Custom Diffusion \cite{kumari2023multi} also fall short in achieving attribute-focused customization. Furthermore, the reliance on multiple reference images for DreamBooth\cite{ruiz2023dreambooth} and Custom Diffusion\cite{kumari2023multi} poses a catastrophic overfitting when fine-tuning on a single reference image. In comparison with Prospect\cite{zhang2023prospect}, our method matches CLIP-T similarity scores closely (0.2822 vs 0.2800, 0.2798 vs 0.2795). Notably, our approach outperforms Prospect in DINO similarity (0.4149 vs 0.3849), Gram matrix distance (0.0791 vs 0.0842), and IoU score (0.4938 vs 0.4656), suggesting superior attribute-focused customization capabilities of our method.


\begin{table*}[]
\caption{Quantitative comparison with respect to the IP-Adapter for appearance and shape attribute customization.}
\label{tab:ip_adapter}
\resizebox{\linewidth}{!}{
\begin{tabular}{c|ccc|ccl}
\hline
Metric                                     & \multicolumn{3}{c|}{Appearance Customization}                                                        & \multicolumn{3}{c}{Shape Customization}                \\ \hline
Method                                     & CLIP-T  $\uparrow$ & DINO Similarity  $\uparrow$ & Gram Matrics $\downarrow$  & CLIP-T  $\uparrow$ & IoU Score $\uparrow$  \\ \hline
IP-Adapter\cite{ye2023ip}                   & 0.2578            & \textbf{0.5157}             & \textbf{0.0449}           & 0.2747              & 0.4157               \\ \hline
Ours                                       & \textbf{0.2822}    & 0.4149                      & 0.0791                     & \textbf{0.2798}    & \textbf{0.4938}      \\ \hline
\end{tabular}
}
\end{table*}

\begin{figure}
    \centering
    \includegraphics[width=1.0\textwidth]{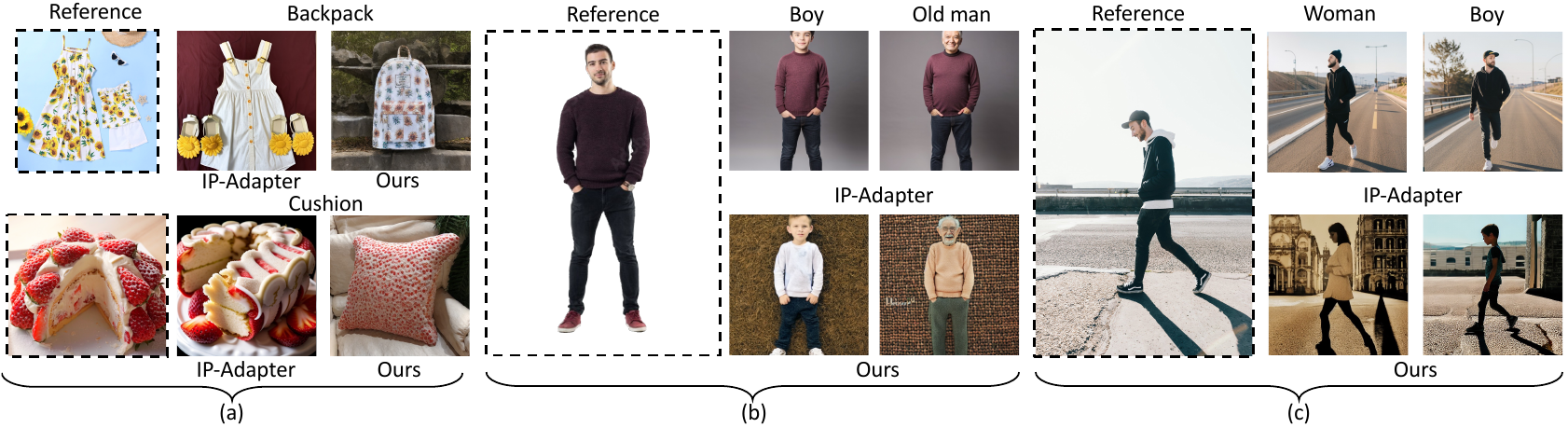}
    \caption{The comparison of generation results between IP-Adapter and our method. Sub-figure (a) showcases the comparison of appearance customization results, while sub-figures (b) and (c) present the comparison of shape customization results.}
    \label{fig:ipadapter}
\end{figure}

\textbf{Comparison with IP-Adapter.} As illustrated in Table \ref{tab:ip_adapter}, we compared our approach with IP-Adapter in terms of appearance and shape customization. While IP-Adapter achieves higher DINO similarity and lower Gram matrix distance, this does not necessarily indicate superiority in customizing appearance attributes. As depicted in Figure \ref{fig:ipadapter} (a), IP-Adapter tends to prioritize learning the content of reference images over their appearance attributes, resulting in images that closely resemble the reference but may not align well with the content described in the text prompt. Conversely, our approach accurately identifies and learns appearance attributes while generating images aligned with specified textual conditions. For shape customization, our method achieves higher IoU and CLIP-T scores, indicating superior shape attribute customization capabilities. Additionally, the IP-Adapter faces the following issues with shape customization: 1) poor decoupling between shape and appearance (see Figure \ref{fig:ipadapter} (b)), 2) excessive focus on identity information, resulting in content generation not being controlled by text (see Figure \ref{fig:ipadapter} (c)). Our method, however, does not suffer from these issues. This demonstrates that our method can achieve high-quality attribute-focused customization results without compromising the controllability of the text.



\begin{table}[]
\caption{Quantitative comparisons with ControlNet (for shape customization) and Prospect (for style customization).}
\label{tab:controlnet}
\resizebox{\linewidth}{!}{%
\begin{tabular}{c|cc|ccc}
\hline
Metric                                                         & \multicolumn{2}{c|}{Shape Customization}                          & \multicolumn{3}{c}{Style Customization}                      \\ \hline
Method                                        & CLIP-T  $\uparrow$ & IoU Score $\uparrow$ & Method                           & Gram matrics $\downarrow$ & CLIP-T $\uparrow$  \\ \hline
ControlNet-Canny\cite{zhang2023adding}        & 0.2688             & \textbf{0.7275}      & Prospect\cite{zhang2023prospect} & 0.0611                    & \textbf{0.3081}            \\ \hline
ControlNet-Segmentation\cite{zhang2023adding} & 0.2788             & 0.7193               & Ours                             & \textbf{0.0340}           & 0.3042             \\ \hline
Ours                                          & \textbf{0.2798}    & 0.4938               & ---                              & ---                       & ---                \\ \hline
\end{tabular}
}
\end{table}

\begin{figure}
    \centering
    \includegraphics[width=1.0\textwidth]{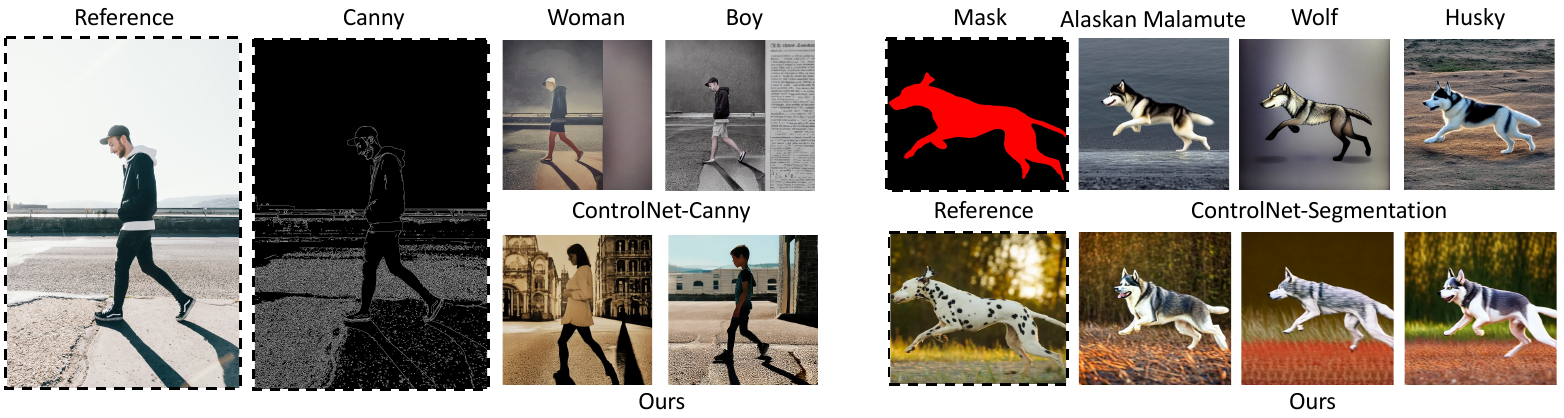}
    \caption{The comparison of shape customization results between ControlNet and Ours.}
    \label{fig:controlnet_compare}
\end{figure}

\textbf{Comparison with ControlNet.} As shown in Table \ref{tab:controlnet} (left), both ControlNet-Canny and ControlNet-Segmentation achieve IoU scores exceeding 0.7. However, this alone cannot demonstrate that the shape customization capability of ControlNet is superior to ours. As illustrated in Figure \ref{fig:controlnet_compare} (left), while the generation results of ControlNet-Canny are generally consistent with the shape of the reference image, they fail to correspond to the target text. This is because the canny prompt retains many shape-irrelevant details, such as clothing and hats. For Figure \ref{fig:controlnet_compare} (right), using a mask as a prompt fails to accurately represent spatial relationships, resulting in generated images that lack realism. For example, the front legs of the dog overlap in the mask, leading to incorrect spatial relationships in the generated content. Our method does not suffer from these issues. Our method can accurately identify shape attributes without relying on any external cues. Moreover, compared to ControlNet, we only require a single image as input, while utilizing a lightweight hypernetwork for fine-tuning, making training more efficient and capable of generating high-quality content. This demonstrates our method has an exemplary balance between efficiency and output quality.

\noindent{\textbf{Evaluation in terms of Style Customization.}} As for the task of style customization, we opt for comparison with Prospect. For each style reference image, we utilize multiple textual prompts to generate a variety of images, resulting in a total of 99 images generated for each method. As shown in Table \ref{tab:controlnet} (right), compared to Prospect, our approach has achieved comparable CLIP-T scores, while also attaining lower Gram matrix distances. This indicates that our method can achieve superior style customization performance without compromising text controllability.

\noindent{\textbf{User Study.}} We conduct 30 experimental studies for the evaluation of appearance, shape, and style attribute customization, respectively. Participants are shown a series of reference images, with images generated from all methods corresponding to each reference image, along with text descriptions for generation. For appearance customization, users select the image that best matches the reference in appearance  and is consistent with the text. For shape customization, users choose the image most visually similar in shape to the primary foreground content of the reference, aligned best with text, and exhibited realistic content. In style customization, users select the image that is closest in visual style to the reference and aligned best with the text. We can observe that our method achieved the highest user study scores compared to other methods in shape, appearance, and style customization. Thus, our method exhibits better performance in human preference for attribute-focused customization generation.

\begin{figure*}
    \centering
    \includegraphics[width=1.0\textwidth]{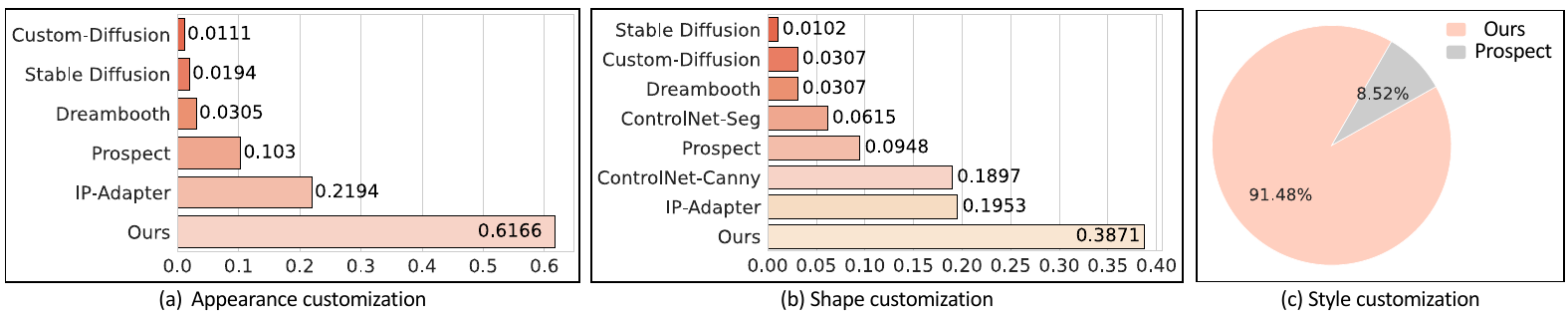}
    \caption{Human preference ratio for different methods on attribute-focused T2I customization.}
    \label{fig:user_study}
\end{figure*}

\vspace{-1cm}

\subsection{Qualitative Experiments}

\begin{figure*}
    \centering
    \includegraphics[width=1.0\textwidth]{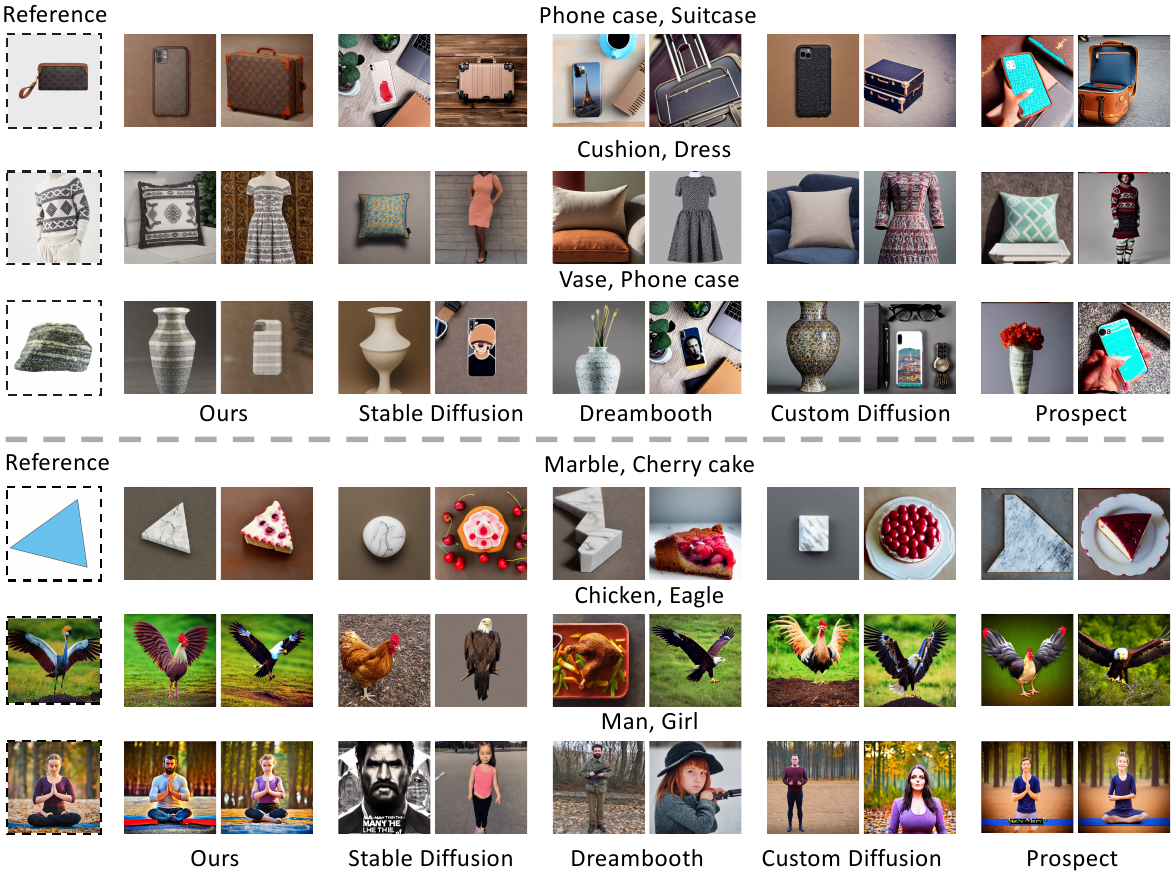}
    \caption{Qualitative comparisons with respect to existing methods for appearance and shape customization. Above dashed line: appearance customization. Below dashed line: shape customization.}
    \label{fig:compare}
\end{figure*}


\begin{figure*}
    \centering
    \includegraphics[width=1.0\textwidth]{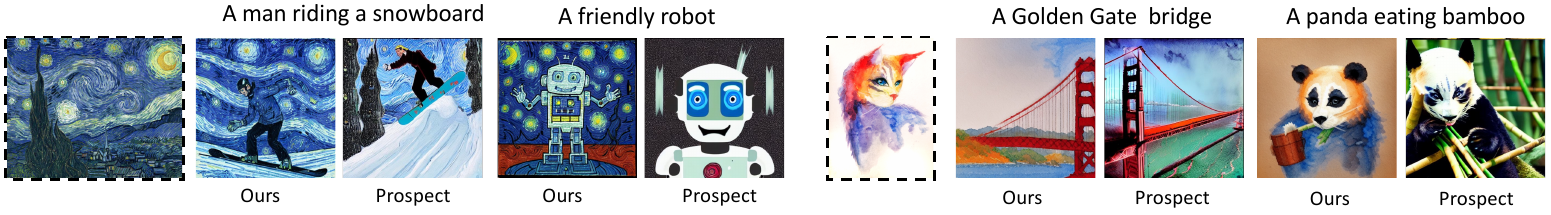}
    \caption{Qualitative comparisons with respect to Prospect for style customization.}
    \label{fig:compare_style}
\end{figure*}

\begin{figure}
    \centering
    \includegraphics[width=1.0\textwidth]{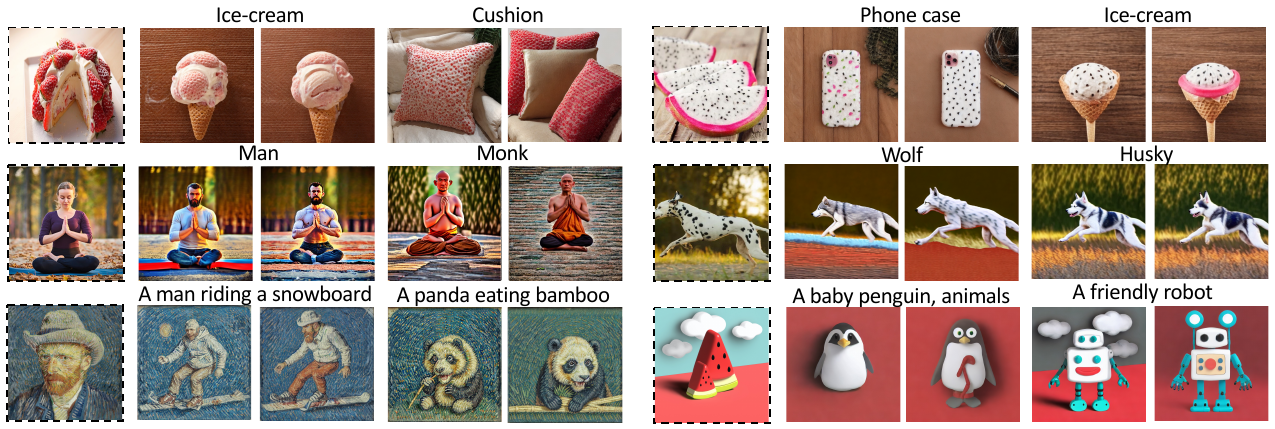}
    \caption{The diverse generation results of our method. Top: appearance customization. Middle: shape customization. Bottom: style customization.}
    \label{fig:diversity}
\end{figure}

\textbf{Qualitative Comparison.} As shown in Figure \ref{fig:compare}, we find that Stable Diffusion\cite{rombach2022high} struggles to recognize and learn shape and appearance attributes, only capable of generating content aligned with textual descriptions. Meanwhile, Dreambooth\cite{ruiz2023dreambooth} and Custom Diffusion\cite{kumari2023multi} exhibit less precise attribute recognition. Prospect\cite{zhang2023prospect} achieves some degree of attribute-aware customization, but is limited to generating categories similar to the reference image, such as woman-to-man or bird-to-chicken. In contrast, our method can accurately identify target attributes while also generating high-quality, cross-domain, text-controlled attribute-focused customization results. The qualitative comparison to IP-Adapter and ControlNet is depicted in Figure \ref{fig:ipadapter} and \ref{fig:controlnet_compare}. Our method, compared to ControlNet and IP-Adapter, maintains text controllability while generating high-quality results in appearance and shape attribute customization. We also present qualitative results of style customization in Figure \ref{fig:compare_style}. Prospect suffers from inadequate decoupling of style attributes due to issues with multiple attributes sharing a sampling stage, resulting in incomplete separation of style attributes (e.g., Prospect's generated panda retains both content and style information from the reference image). In contrast, our method achieves precise identification of style attributes while also producing high-quality customization results.

\textbf{Diversity.} 
We present a range of diverse generation results to validate the diversity generation capability of our method. As shown in Figure \ref{fig:diversity}, our method achieves various results for attribute-focused customization, demonstrating its exceptional diversity generation capability.

\begin{figure}
    \centering
    \includegraphics[width=1.0\textwidth]{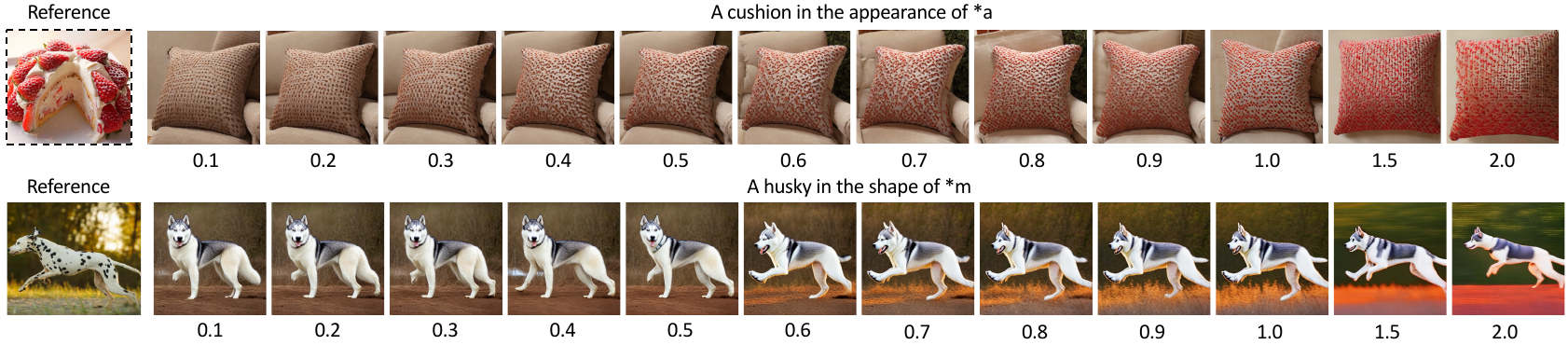}
    \caption{ The generation results of our method under different $\lambda$ settings.}
    \label{fig:lambda}
\end{figure}

\textbf{Adjustable Attribute Customization Intensity.} Our method provides flexible control over attribute customization intensity via the weight coefficient $\lambda$. In Figure \ref{fig:lambda}, we illustrate the impact of different $\lambda$ values on customization outcomes. Increasing $\lambda$ results in closer alignment of the attributes between the generated and the target images. This capability enables users to finely adjust customization levels by selecting suitable weights, a feature not found in alternative approaches.

\vspace{-1cm}
\begin{figure}
    \centering
    \includegraphics[width=1.0\textwidth]{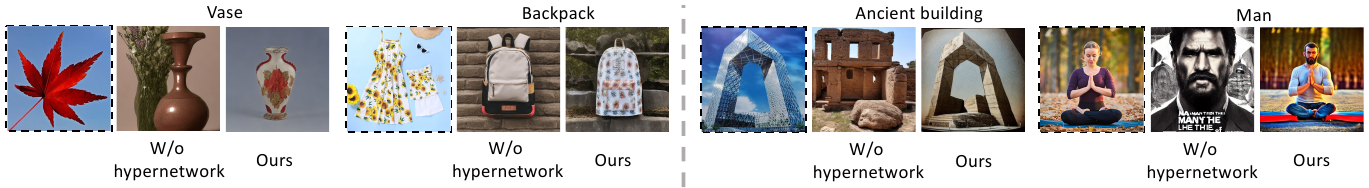}
    \caption{The ablation results of hypernetwork-driven fine-tuning strategy.}
    \label{fig:ablation}
\end{figure}

\subsection{Ablation Study}
We conduct ablation experiments with the hypernetwork-driven fine-tuning strategy, as shown in Figure \ref{fig:ablation}. Without utilizing the hypernetwork, the network struggles to learn complex image data and often generates content unrelated to the target attributes. In contrast, when the hypernetwork was employed, the network successfully learned complex image data, generating content such as a backpack resembling a dress or ancient architecture resembling modern architecture. These results highlight the effectiveness of the hypernetwork-driven fine-tuning strategy.

\section{Conclusions, Limitations and Future Work}
\label{sec:conclusion}

We introduce OSTAF as a novel approach for one-shot, attribute-focused text-to-image (T2I) personalization. It stands out from existing subject-driven or attribute-aware personalization techniques by focusing on the precise identification, representation, and replication of specific attributes from a reference image. We conduct a detailed analysis of how the U-net encoder and decoder process various attributes, which motivates the development of an efficient hypernetwork-enhanced and attribute-focused fine-tuning strategy. This allows for rapid and accurate attribute-specific customization of pre-trained T2I models. Through comprehensive evaluation, our method outperforms existing solutions in the domain of attribute-focused text-to-image customization.
While our method is efficient by only fine-tuning with a single reference image, the tuning time required is on par with that of Prospect and P+, suggesting room for improvement. Our future work will aim to accelerate the fine-tuning process and extend our technique to video content, enabling more dynamic and detailed attribute customization.

\clearpage  

%
%
\bibliographystyle{splncs04}
\bibliography{main}

\clearpage
\section*{Supplementary Material}

To better facilitate understanding of our work, we provide more comprehensive details in this supplementary material. Firstly, we  introduce additional training details regarding our model. Secondly, we present the datasets we have collected. Thirdly, we elaborate on the specifics of the user study conducted. Finally, we present more qualitative results of our method. 

\section*{Training Details}

\begin{table}[]
\caption{Hyperparameters setting for attribute-focused customization.}
\label{tab:hyperparameters}
\centering

\begin{tabular}{cccc}
\hline
\multirow{2}{*}{Hyperparameters} & \multicolumn{3}{c}{Attribute}                 \\ \cline{2-4} 
                                  & Appearance    & Shape         & Style         \\ \hline
CLIP version                      & ViT-B-32      & ViT-B-32      & ViT-B-32      \\
Domain class name                 & \$class\_name & \$class\_name & \$class\_name \\
Attribute word                    & appearance    & shape         & style         \\
Learning rate                     & 1e-6          & 1e-6          & 1e-6          \\
Max train steps                   & 3000          & 3000          & 3000          \\
Mixed precision                   & fp16          & fp16          & fp16          \\
Placeholder token                 & *a            & *m            & *s            \\
Stable Diffusion version          & sd-1.4        & sd-1.4        & sd-1.4        \\
Resolution                        & 512           & 512           & 512           \\
Batch size                        & 1             & 1             & 1             \\
Gradient accumulation steps       & 2             & 2             & 2             \\
Use 8bit adam                    & true          & true          & true          \\
Random seed                       & 42            & 42            & 42            \\ \hline
\end{tabular}
\end{table}

As illustrated in Table \ref{tab:hyperparameters}, we present the hyperparameter settings for training appearance customization, shape customization and style customization, respectively. In these configurations, \texttt{\$class\_name} denotes the category information of the current image content. For each attribute, we utilize different placeholder tokens, namely *a, *m and *s. Their initialization process will be elaborated upon in the subsequent sections. During the training phase, we leverage mixed-precision models to expedite the training process. The maximum number of training iterations is set to 3,000 (typically, 1,000 iterations are sufficient to attain satisfactory results) to ensure comprehensive learning of the attributes. Furthermore, we employ the gradient accumulation technique, updating the gradients every 2 iterations to enhance training efficiency.

\noindent{\textbf{Placeholder Tokens Initialization.}} We utilize multimodal fusion embeddings to initialize the placeholders *a, *m  *s. Specifically, we employ the CLIP image encoder to extract visual features from the reference image. These visual features are then fused with the CLIP text embeddings of the attribute words,i.e., \textit{shape}, \textit{appearance} and \textit{style}, enabling the embedding initialization.

\section*{Attribute Benchmark}

\begin{figure*}
    \centering
    \includegraphics[width=1.0\textwidth]{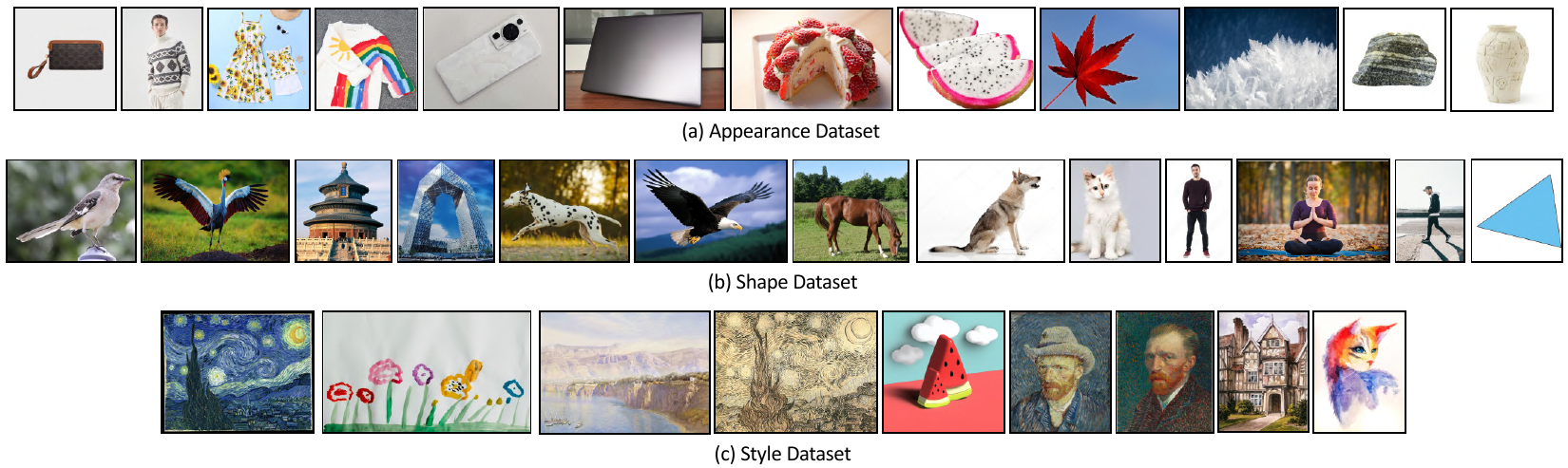}
    \caption{All reference images of appearance dataset, shape dataset and style dataset.}
    \label{fig:dataset}
\end{figure*}

As illustrated in Figure \ref{fig:dataset}, the Attribute Benchmark comprises three distinct sub-datasets: the appearance dataset consisting of 12 images, the shape dataset containing 13 images and the style dataset encompassing 9 images. The categories of data we collected are highly diverse. Taking the shape dataset as an example, the collected images span a wide range of categories, including animals, buildings, humans and basic geometric shapes. Furthermore, in the appearance dataset, we not only encompass common apparel images but also gather images of food, natural objects (such as ice and rocks) and more. Finally, we focus on nine distinct stylistic genres for the style dataset.
In summary, the Attribute Benchmark, with its comprehensive and diverse collection of images across various categories, serves as a valuable dataset for facilitating the training and evaluation of attribute-focused image customization models, enabling further exploration of this challenging task.

\section*{User Study}

\begin{figure*}
    \centering
    \includegraphics[width=1.0\textwidth]{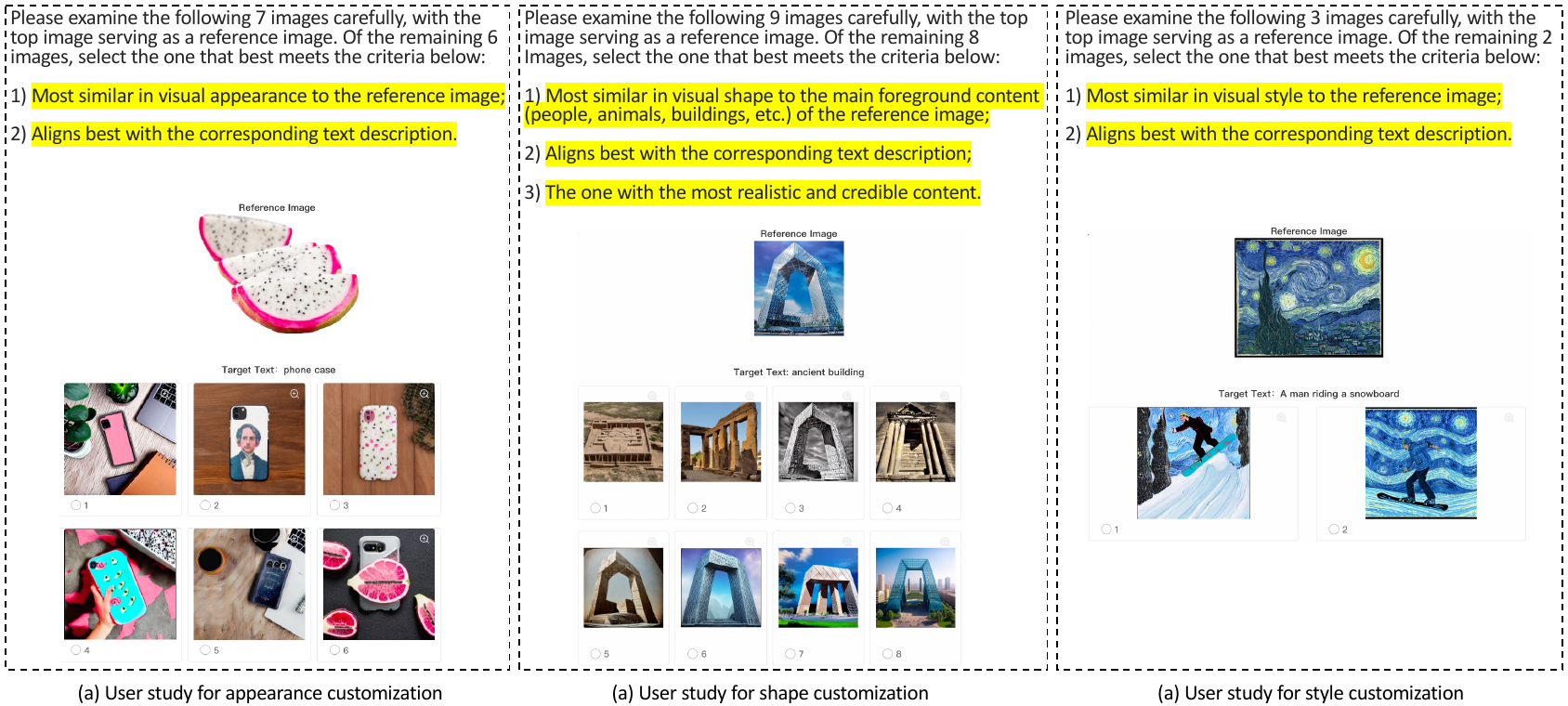}
    \caption{Examples of the user study for attribute-focused image customization}
    \label{fig:gui}
\end{figure*}

Figure \ref{fig:gui} illustrates the details of the user study we conducted, which is divided into three parts corresponding to the user study evaluation interfaces for appearance, shape and style attribute customization.

Each part provides a reference image along with a target text description, and users are asked to select the best match from multiple alternative images. The criteria for selection vary by the specific task but generally include visual similarity, the relevance of the text description and the fidelity of the image content, as highlighted in the yellow text.

For instance, in the appearance customization task, the reference image is a dragon fruit, and the target description is "phone case." Users should choose one image from six options, so that the selected image most accurately reflects the appearance of the dragon fruit while also closely resembling a phone case. In the shape customization task, users are asked to select the image that most closely matches the shape of the reference image, based on the target description "ancient building", from a selection of eight images generated by different methods.
Lastly, in the style customization section, users are asked to determine which one of the two options best captures the style of the reference image and aligns with the text description "A man riding a snowboard."
The user study evaluation process for the remaining reference images follows a similar approach as depicted in Figure \ref{fig:gui}.

\section*{More Qualitative Results}

\begin{figure*}
    \centering
    \includegraphics[width=1.0\textwidth]{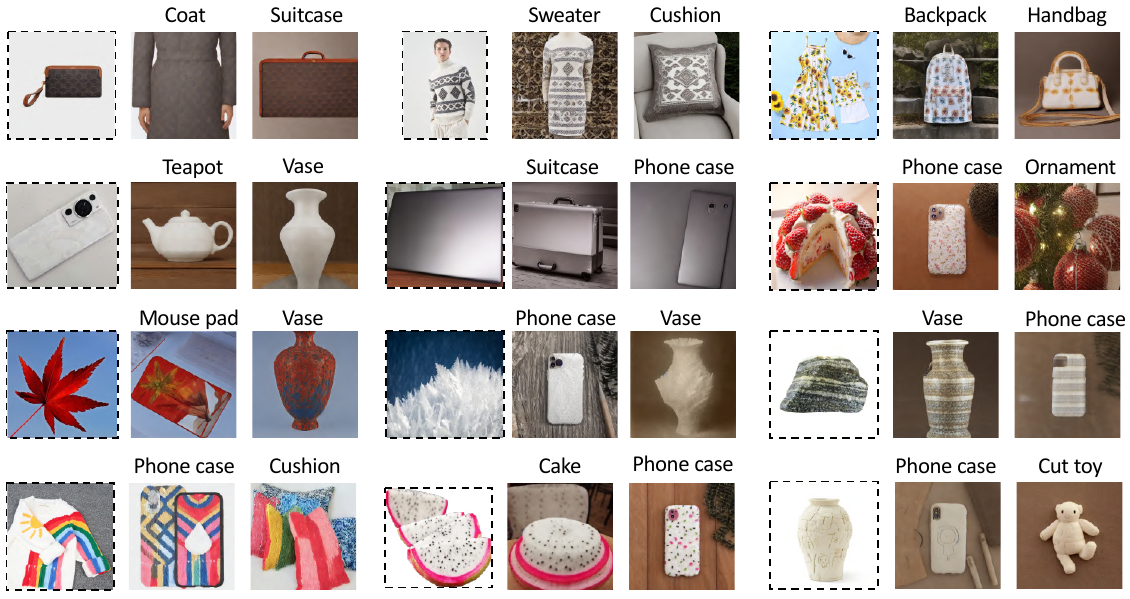}
    \caption{ More appearance customization results. The reference image is presented within the dashed rectangular frame.}
    \label{fig:app_more}
\end{figure*}

\begin{figure*}
    \centering
    \includegraphics[width=1.0\textwidth]{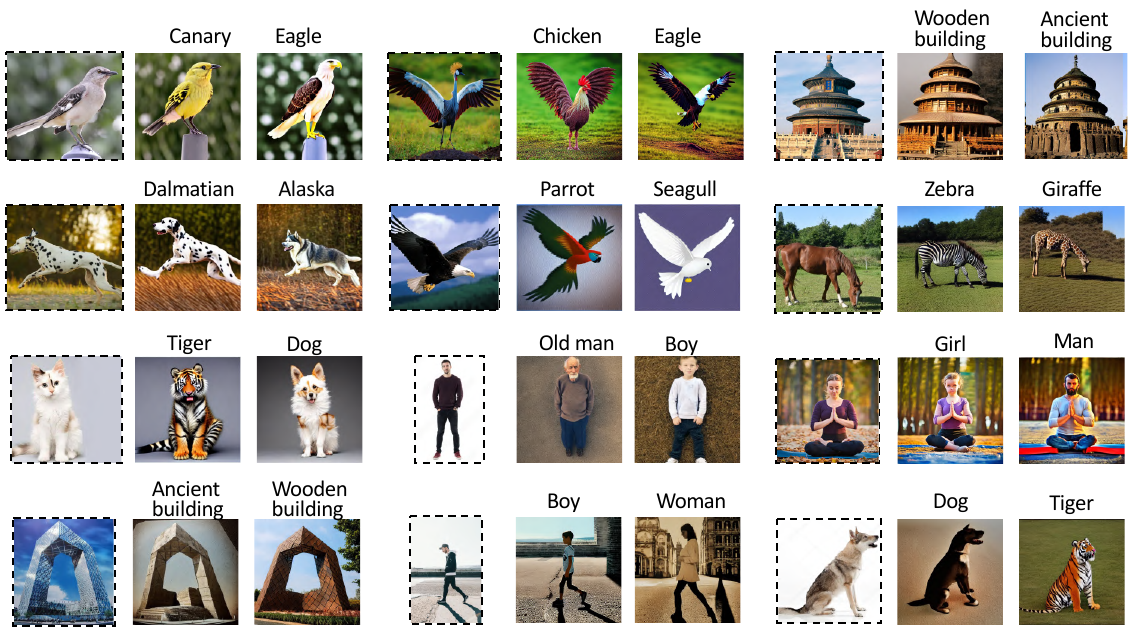}
    \caption{ More shape customization results. The reference image is presented within the dashed rectangular frame.}
    \label{fig:shape_more}
\end{figure*}

As shown in Figures \ref{fig:app_more} and \ref{fig:shape_more}, we have provided more attribute-focused customization results. For each given reference image, we have generated two new images. From the results, our method achieves more precise attribute recognition and high-quality customized generation.

\end{document}